%% file: main.tex
\documentclass[10pt,twocolumn,letterpaper]{article}

\usepackage{cvpr}              

\input{preamble}

\definecolor{cvprblue}{rgb}{0.21,0.49,0.74}
\usepackage[pagebackref,breaklinks,colorlinks,allcolors=cvprblue]{hyperref}
\usepackage[ruled,vlined]{algorithm2e}
\usepackage{pifont}
\usepackage{multirow}
\usepackage[accsupp]{axessibility}  
\def\paperID{5411} 
\def\confName{CVPR}
\def\confYear{2026}

\title{S2D: Sparse to Dense Lifting for 3D Reconstruction with Minimal Inputs}

\author{
Yuzhou Ji$^{1}$, Qijian Tian$^{1}$, He Zhu$^{1}$, Xiaoqi Jiang$^{4}$, Guangzhi Cao$^{4}$, Lizhuang Ma$^{1}$, Yuan Xie$^{2}$, Xin Tan$^{2,3}$\footnotemark[1]\\
$^{1}$Shanghai Jiao Tong University\quad
$^{2}$East China Normal University\quad \\
$^{3}$Shanghai Artificial Intelligence Laboratory\quad
$^{4}$Chery Automobile\\
{\tt\small jiyuzhou@sjtu.edu.cn}\quad {\tt\small xtan@cs.ecnu.edu.cn}
}

\begin{document}

\renewcommand{\thefootnote}{\fnsymbol{footnote}}

\twocolumn[{%
\renewcommand\twocolumn[1][]{#1}%

\maketitle

\begin{center}
  \includegraphics[width=\linewidth]{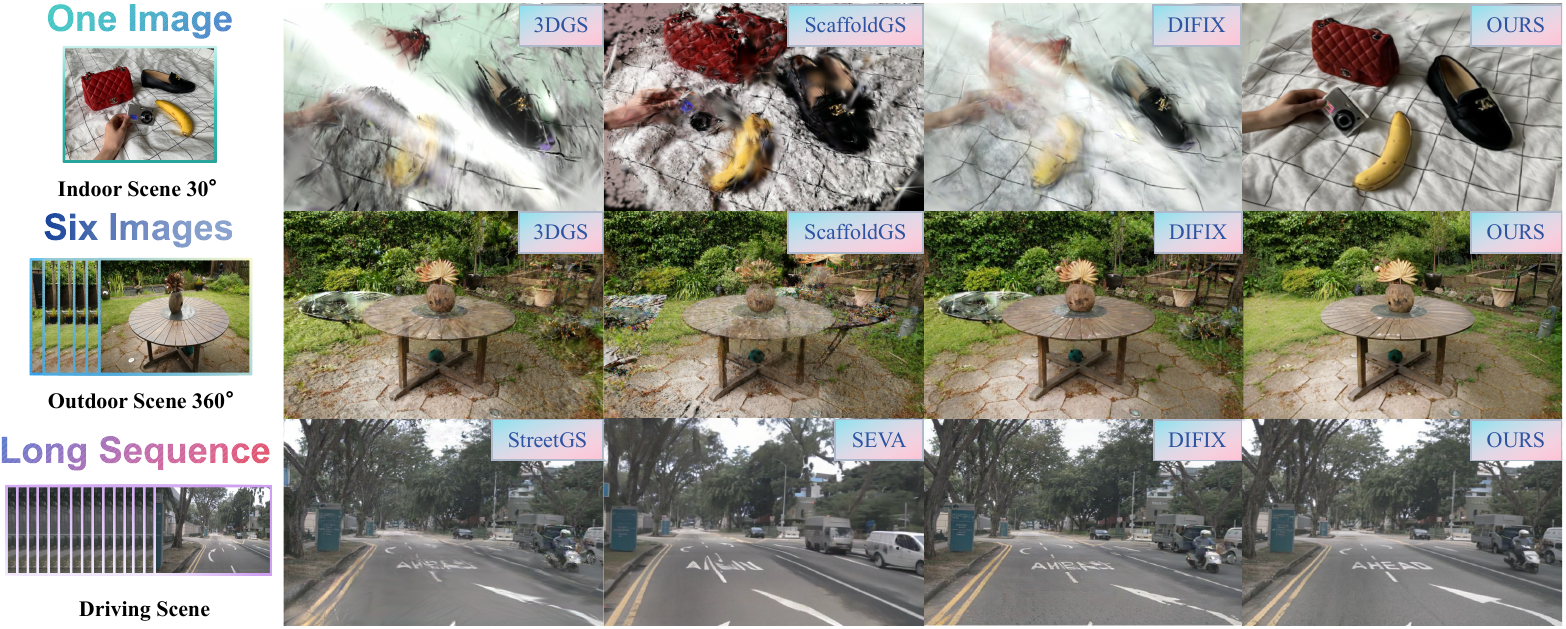}
  \captionof{figure}{\textbf{We demonstrate S2D in different situations.} S2D supports reconstruction with inputs of different density. The 3DGS scene reconstructed through S2D is free of severe artifacts that occur in traditional reconstruction under sparse inputs. S2D also outperforms state-of-the-art scene enhancer (DIFIX) and direct video generation conditioned on camera poses (SEVA).}
  \label{fig:teaser}
\end{center}
\vspace{0.3\baselineskip}
}]

\footnotetext[1]{Corresponding author.}

\input{sec/0_abstract}    
\input{sec/1_intro}
\input{sec/2_related}
\input{sec/3_method}
\input{sec/4_experiment}

\input{sec/5_ablation}
\input{sec/6_conclusion}
\input{sec/7_ack}
{
    \small
    \bibliographystyle{ieeenat_fullname}
    \bibliography{main}
}

\input{sec/X_suppl}

\end{document}

%% file: preamble.tex
\newcommand{\TODO}[1]{\textbf{\color{red}[TODO: #1]}}
\renewcommand{\TODO}[1]{}



%% file: sec/0_abstract.tex
\begin{abstract}
Explicit 3D representations have already become an essential medium for 3D simulation and understanding.
However, the most commonly used point cloud and 3D Gaussian Splatting (3DGS) each suffer from non-photorealistic rendering and significant degradation under sparse inputs.
In this paper, we introduce Sparse to Dense lifting (S2D), a novel pipeline that bridges the two representations and achieves high-quality 3DGS reconstruction with minimal inputs.
Specifically, the S2D lifting is two-fold.
We first present an efficient one-step diffusion model that lifts sparse point cloud for high-fidelity image artifact fixing.
Meanwhile, to reconstruct 3D consistent scenes, we also design a corresponding reconstruction strategy with random sample drop and weighted gradient for robust model fitting from sparse input views to dense novel views.
Extensive experiments show that S2D achieves the best consistency in generating novel view guidance and first-tier sparse view reconstruction quality under different input sparsity.
By reconstructing stable scenes with the least possible captures among existing methods, S2D enables minimal input requirements for 3DGS applications.
Project page at \url{https://george-attano.github.io/S2D}.
\end{abstract}

%% file: sec/1_intro.tex
\section{Introduction}
\label{sec:intro}

While pure vision language models still show certain limitation in spatial understanding due to 2D inputs, 3D Gaussian Splatting (3DGS) \cite{kerbl20233d} has long been playing an important role in autonomous driving and embodied AI simulation.
However, the rendering quality of 3DGS degrades significantly when the viewing angle deviates from input poses, which requires a large amount of input to maintain low view interpolation distance.
In real-life circumstances, it is unrealistic to always ensure dense input, not to even mention the corresponding computational cost.
Therefore, the practical application based on 3DGS has been severely hindered.

To get rid of the long-lasting constraint of input requirements hovering upon 3DGS, the ultimate goal is to achieve photorealistic reconstruction with minimal inputs.
With this aim, we first define the most demanding sparse input setting to be acquiring the same viewing range as in point cloud reconstruction.
For example, with only one input image, the reconstructed scene can be viewed within $30^\circ$ spiral angles, and can stably provide more than $180^\circ$ view range with less than 10 input images.
In such a setting, traditional reconstruction methods are corrupted as shown in Figure \ref{fig:teaser}.

To solve this problem, some try to build a feed-forward models \cite{charatan2024pixelsplat, szymanowicz2024splatter, chen2024mvsplat, Tian2025drive, tang2024lgm, xu2024grm, zhang2024gs, xu2025depthsplat, jiang2025anysplat, wang2025volsplat} that directly predicts gaussian attributes, but still creates much artifacts in this scenario (as in Figure \ref{fig:qualitative}).
While object-centric methods \cite{zou2023triplane, tang2023dreamgaussian, yang2024gaussianobject} can not be applied to general reconstruction, others try to directly generate novel views through diffusion models conditioned on camera extrinsics or sparse points \cite{voleti2024sv3d, wang2024motionctrl, yu2024viewcrafter, watson2025controllingspacetimediffusion, zhou2025stable}, but fail to maintain 3D consistency and generation fidelity, which also require high time consumption.
Instead, the recent DIFIX \cite{wu2025difix3d+} builds a generative fixer that can remove novel view artifacts, which supports scene reconstruction with relatively sparser inputs.
However, DIFIX is built upon small view deviation and only operates on mild artifacts.
Meanwhile, DIFIX overlooks the gap between novel guidance and real input and directly conduct reconstruction, causing severe 3D inconsistency.
In our settings, DIFIX completely fails in sparse inputs, leaving this challenge far from solved. 

While 3D consistency and guidance have become a common obstacle in the methods mentioned above, the latest vision foundation models (VFM) \cite{wang2025vggt, wang2025pi3, keetha2025mapanything} have revolutionized 3D prediction, which can conduct dense point cloud reconstruction instantly.
Although point cloud requires few input images and is naturally view independent, its rendering is far from photorealistic, with noticeable noise due to aliasing and cumulative error, constraining its application in image-level guidance.

To address the aforementioned challenges, in this paper, we present Sparse to Dense lifting (S2D), a flexible framework that provides high-quality 3DGS reconstruction with extremely sparse inputs.
Specifically, we first contribute a strong yet efficient one-step diffusion model that lifts the sparse point cloud for high-fidelity image artifact fixing. 
With pixel-level details provided by the input view and structural guidance in point cloud rendering, our fixing model is able to handle extreme artifacts while still providing excellent consistency across views.
For better fitting the 3DGS optimization under sparse inputs and dense fixed guidance, we also propose a corresponding reconstruction strategy with random sample drop and weighted gradient.
Extensive experiments show that S2D achieves the best consistency in generating novel view guidance and first-tier sparse view reconstruction quality among various methods.

Notably, S2D is not fixed in the input number and supports any input density, which is suitable for general reconstruction enhancement.

In summary, this work has the following contributions:
\begin{itemize}
    \item We propose S2D, a flexible framework for baseline 3DGS methods to conduct sparse-view reconstruction, supporting view extrapolation and larger view interpolation.
    \item We present a strong yet efficient artifact-fixing model that can take both input view and novel view point cloud rendering as guidance, achieving first-tier quality in removing image artifacts.
    \item We design a reconstruction strategy for better model fitting and 3D consistency under sparse input and dense novel guidance.
\end{itemize}

%% file: sec/2_related.tex
\section{Related Work}
\subsection{Gaussian Splatting}
After first introduction, 3D Gaussian Splatting (3DGS) \cite{kerbl20233d} has expanded upon largely with further innovations \cite{sun2025splatflow, lu2025bard, peng2025gaussian, wu2024recent}.
Many contributions \cite{liu2025citygaussian, chen2024mvsplat, xie2025envgs, Zheng2025WildGS} underscore the ongoing enhancements and versatility of 3DGS in handling increasingly complex rendering tasks.
Despite the advantages of 3D Gaussian Splatting in rendering speed and quality, the results degrade significantly with the reduction of input training views, limiting the applicable scenarios.

To this end, feed-forward 3DGS models \cite{charatan2024pixelsplat, szymanowicz2024splatter, chen2024mvsplat, Tian2025drive, tang2024lgm, xu2024grm, zhang2024gs, xu2025depthsplat, jiang2025anysplat, wang2025volsplat, ye2024noposplat} have emerged to handle reconstruction with fewer inputs.
In particular, pixelSplat \cite{charatan2024pixelsplat} and Splatter Image \cite{szymanowicz2024splatter} predict Gaussian attributes from image features, while MVSplat \cite{chen2024mvsplat} encodes the feature matching information with cost volumes.
Depthsplat \cite{xu2025depthsplat} further integrates monocular features from pre-trained monocular depth models and achieves more robust depth prediction and 3D Gaussian reconstruction.

Although these methods improve quality and efficiency in sparse-view reconstruction, the lack of explicit 3D supervision limits generalization of pre-trained models and is still unable to handle extreme inputs.

\subsection{Universal 3D reconstruction}
Feed-forward models has advanced even more significantly in point cloud reconstruction while predicting multiple 3D attributes.
The early DUSt3R \cite{wang2024dust3r} and MASt3R \cite{leroy2024grounding} predict a coupled scene representation but require further post-processing.
The following works expand them towards more pipelines \cite{duisterhof2025mast3r, murai2025mast3r, elflein2025light3r, pataki2025mp} and support extra input views \cite{cabon2025must3r, wang20243d, wang2025continuous}, but still provide limited quality compared to traditional optimization.

Recently, multi-view models such as VGGT \cite{wang2025vggt} have made excellent progress in 3D prediction, with many valuable efforts in more accurate localization \cite{tang2024mv, reloc3r} and longer sequences \cite{Yang_2025_Fast3R, deng2025vggtlongchunkitloop}.
Notably, $\pi^3$ \cite{wang2025pi3} employs a fully permutation-equivariant architecture and achieves higher robustness.
Later, MapAnything \cite{keetha2025mapanything} further enables a broad range of 3D vision tasks in a single feed-forward pass, pushing such models to real-world applications.
While these methods do not directly support photorealistic view synthesis, we utilize them as strong structural consistent guidance for image level lifting.

\subsection{Reconstruction with diffusion prior}
Advancements in image and video diffusion models \cite{rombach2022high, blattmann2023stable, podell2023sdxl, wang2024egovid, yang2024cogvideox} have achieved highly controlled generation, making it possible to leverage diffusion prior as photorealistic guidance for scene reconstruction.
In autonomous driving simulation, it has already been a popular solution to distill customized video generation to dynamic scene reconstruction \cite{zhao2025drivedreamer4d, wang2024drivedreamer, zhao2025drivedreamer, yan2025streetcrafter, ni2025recondreamer, zhao2025recondreamer++}.
Concerning common static scenarios, diffusion-based novel view synthesis methods have also evolved from object-level to scene-level \cite{liu2023zero1to3, zeronvs, wu2023reconfusion, gao2024cat3d, voleti2024sv3d, wang2024motionctrl, yu2024viewcrafter, watson2025controllingspacetimediffusion, zhou2025stable, fan2025freesim, wei2025gsfix3ddiffusionguidedrepairnovel, diffusiongs, wang2025diffusion}, but their generalization ability, time consumption, and 3D consistency are still far from practical applications.

The recent DIFIX3D+ \cite{wu2025difix3d+} trains an image diffusion model that takes in surround view reference to restore novel view artifacts as extra guidance, achieving efficient reconstruction with extrapolated views.
However, Difix significantly degrades when faced with extreme artifacts under large view deviation, hindering its further applications.

In contrast, S2D gets rid of these limitations and provides efficient and consistent sparse-view reconstruction with structurally guided artifact fixing and lifting.

%% file: sec/3_method.tex
\section{Method}
\begin{figure*}[!t]\centering
  \includegraphics[width=1\linewidth]{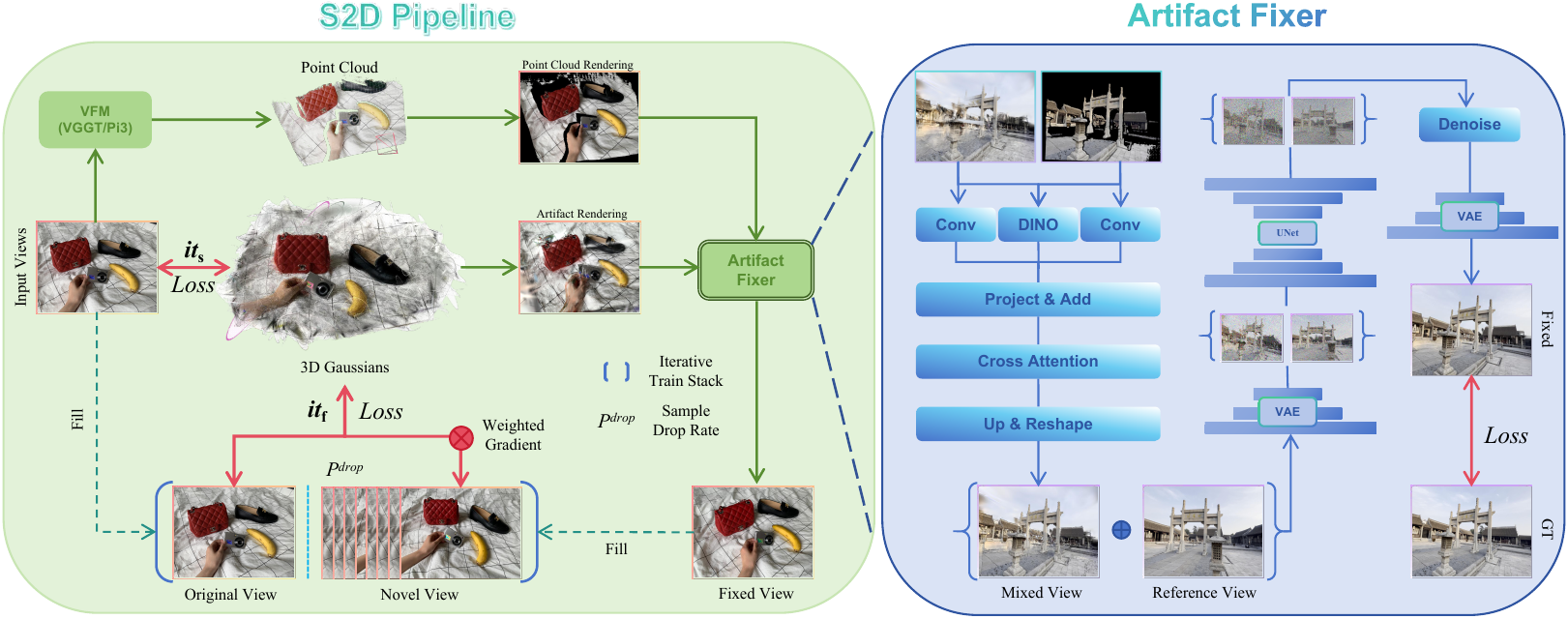}
  \caption{S2D reconstruction pipeline and model architecture of artifact fixer.}
  \label{fig:pipeline}
\end{figure*}
\subsection{Overview}
The general pipeline of S2D is shown in Figure \ref{fig:pipeline} Left.
Given any number of input views and novel cameras, we firstly generate the point cloud using VFM \cite{wang2025vggt, keetha2025mapanything, wang2025pi3} and render views on novel cameras.
Then we initialize the 3DGS scene on input views until sample iteration $it_s$.
At $it_s$, a single round of rendering on all novel cameras is conducted.
The resulting novel view images containing artifacts are then sent through our artifact fixer (detailed structure in Figure \ref{fig:pipeline} Right) along with a nearby input view as reference and the corresponding point cloud rendering.
After removing artifacts, we continue to optimize the scene with mixed supervision from input views and novel view results until final iteration $it_e$, where random sample drop and weighted gradient are employed.
Finally, we obtain the consistent 3DGS scene with largely expanded view range.

\subsection{Initialization}
Given any input views, we conduct simple initialization.
We sent the original views through VFM ($\pi^3$ \cite{wang2025pi3} with default setting in practice) to acquire the sparse point cloud of the scene.
A set of novel cameras can be automatically generated (\eg through interpolation or preset extrapolation offsets) or manually set according to needs.
We render views of the point cloud on these views as later guidance.

\subsection{Novel view artifact fixer}
\label{sec:method_fixer}
While the distillation of video diffusion suffers from high consumption \cite{wu2023reconfusion, ni2025recondreamer, zhao2025recondreamer++, yan2025streetcrafter}, methods based on image enhancement also show incompetence in extreme situations \cite{wu2025difix3d+, fan2025freesim}.
In this work, we build a generative artifact fixer upon one-step diffusion model yet supports strong structural guidance, which provides both high efficiency and quality.

\textbf{One-step diffusion model.}
In Latent Diffusion Models (LDM) \cite{rombach2022high}, the data \( x_t \) is encoded into a latent representation \( z_t \) via the perceptual compression encoder $\mathcal{E}$, and samples from data distribution $p(z)$ is decoded to image space with a single pass through decoder $\mathcal{D}$.
LDM realizes the neural backbone $\epsilon_\theta(o,t)$ as a time-conditional UNet \cite{ronneberger2015u}, with the simplified objective as:
\begin{equation}
    L_{L D M}=\mathbb{E}_{\mathcal{E}(x), \epsilon \sim \mathcal{N}(0,1), t}\left[\left\| \epsilon-\epsilon_{\theta}\left(z_{t}, t\right)\right\| _{2}^{2}\right] .
    \label{eq:ldm}
\end{equation}
In pix2pix-turbo \cite{parmar2024one}, the backbone $\mathcal{E}$, $\mathcal{D}$ and $\epsilon_\theta$ are finetuned at fixed timestep $t_d$ so that image translation can be efficiently conducted through a single denoising step instead of multiple steps.

\textbf{Dual guidance.}
The state-of-the-art method Difix \cite{wu2025difix3d+} uses nearby views as reference for artifact fixing, which mainly provides texture guidance that fixes blurry regions but fails in larger artifacts and corrupted structure.

While point cloud is naturally view independent and structurally consistent, we introduce dual guidance for artifact fixing.
For each novel view rendering, it is fixed with both corresponding point cloud rendering and nearby reference.
Although accurate point cloud can serve as strong hint for details, noise is currently inevitable due to aliasing and cumulative error, thus we design a mixing module to determine the sample of guidance so that only valuable parts should be considered.

The model architecture of our artifact fixer is shown in Figure \ref{fig:pipeline} Right.
Specifically, in the early mixing module, we extract both DINO features and images features of target view and corresponding point cloud guidance.
The features go through project layers and combined.
After cross attention step, the embeds are lifted to generate a mixed input image.
This mixing module is designed to encourage better usage of point cloud guidance ( see ablations in Sec \ref{sec:guidancemixing}).

The mixed image $I^m$ and reference image $I^r$ are encoded through VAE encoder $\mathcal{E}$ to obtain latent representation $z$:
\begin{equation}
    [z^m, z^r] = \mathcal{E}([I^m, I^r]),
    \label{eq:vaeencode}
\end{equation}
and the noise sample $Z$ at fixed timestep $t_d$ is predicted by multi-view time-conditional UNet \cite{wu2025difix3d+}:
\begin{equation}
    [Z^m, Z^r]=\epsilon_{\theta}\left([z^m,z^r], t_d\right) .
    \label{eq:unet}
\end{equation}
We discard $Z^r$ and conduct single step denoise conditioned on $z^m$:
\begin{equation}
    z^d = \frac{\sqrt{\xi_{t-1}} \cdot \eta_t}{\eta_{t-1}} \cdot Z^m + \frac{\sqrt{\xi_t} \cdot \eta_{t-1}}{\eta_t} \cdot z^m + \sigma_t \cdot \epsilon_t ,
    \label{eq:denoise}
\end{equation}
where $t=t_d$, and finally decode $z^d$ to fixed image $I^{fix}$:
\begin{equation}
    I^{fix} = \mathcal{D}(z^d).
    \label{eq:vaedecode}
\end{equation}

\textbf{Train artifact fixer.}
We first generate paired training data.
We train DL3DV-960-10K scenes on 3DGS on 4000 iterations with 25 to 50 sample intervals to obtain novel view images with different levels of artifacts.
Apart from this sparse reconstruction method, we also apply random perturbations to both the position and rotation attributes of Gaussians at rendering time to generate more artifacts.
The rendering time position $\mathbf{x}_i' \in \mathbb{R}^3$ is generated by:
\begin{equation}
\mathbf{x}_i' = \mathbf{x}_i + \epsilon_i^{\mathbf{x}},
\quad
\epsilon_i^{\mathbf{x}} \sim \mathcal{N}(0, \sigma_x^2 \mathbf{I}_3)
\label{eq:pos_jitter}
\end{equation}
where $\mathbf{x}_i \in \mathbb{R}^3$ is the original 3D coordinate of the $i$-th Gaussian, and $\sigma_x$ controls the magnitude of the positional perturbation ($\sigma_x \in [10^{-4}, 10^{-2}]$).
The rendering time rotation is generated by:
\begin{equation}
\mathbf{q}_i' = \mathbf{q}_{\epsilon_i} \otimes \mathbf{q}_i
\label{eq:rot_jitter}
\end{equation}
The random rotation quaternion $\mathbf{q}_{\epsilon_i}$ is generated from an axis-angle representation:
\begin{equation}
\mathbf{q}_{\epsilon_i} = \left[ \cos\frac{\phi_i}{2}, \, \mathbf{k}_i \sin\frac{\phi_i}{2} \right]
\label{eq:rand_quat}
\end{equation}
where $\mathbf{q}_i \in \mathbb{H}$ is the original unit quaternion, $\mathbf{k}_i \sim \mathcal{U}(\mathbb{S}^2)$ is a random unit vector (rotation axis), $\phi_i \sim \mathcal{U}[-\delta_\phi, \delta_\phi]$ is a random rotation angle ($\delta_\phi \in [5^\circ, 45^\circ]$), $\otimes$ denotes quaternion multiplication (Hamilton product).

We input all views to VFM to align test cameras, but the corresponding point cloud images are only rendered with points reconstructed from the same sample views.

We use pix2pix-turbo \cite{parmar2024one} as the backbone of our artifact fixer, and initialize UNet weights from SD-Turbo \cite{sauer2023adversarialdiffusiondistillation} and VAE weights from Difix \cite{wu2025difix3d+}.
We train the mixing module and LoRA \cite{hu2022lora} adapters applied on VAE and UNet modules.
Loss is computed against fixed images and ground truth images.
We replace the CLIP \cite{radford2021learning} loss in the backbone with DINO feature loss (as in cosine similarity) and add Similarity Index Measure (SSIM), and inherit the original GAN loss.
So the final loss term is:
\begin{equation}
    \mathcal{L} = 0.2\mathcal{L}_{LPIPS} + 0.4\mathcal{L}_2 + 0.25\mathcal{L}_{SSIM} + \mathcal{L}_{GAN} + \mathcal{L}_{DINO}.
    \label{eq:lossfixer}
\end{equation}

\subsection{Reconstruction with novel guidance}
With extremely sparse inputs, it is also a challenge to avoid overfitting on novel views and underfitting on input views.
For example, extrapolated views may have regions that are not exist within the original view range, and inconsistency could occur in parts with rich textures, which also causes severe artifacts.
Previous methods either overlook this problem or simply use smaller weights on novel views \cite{ni2025recondreamer, zhao2025recondreamer++, yan2025streetcrafter}, which introduces an unavoidable optimization bias and fails to account for the inherent gap between input and novel viewpoints.
In this paper, we propose random sample drop and weighted gradients for stable optimization under such circumstances.

\textbf{Random sample drop.}
To mitigate quantitative gap between original views and novel views in the supervision set (\eg a large expansion including 6 input views and 300 novel views), we adopt a probabilistic sampling strategy in training to ensure that original views can provide continuous and sufficient supervision while expanding to novel views. The reconstruction process with is random sample drop depicted in Algorithm \ref{alg:rsd}.

Specifically, given input reference views $V_{ref}$ and novel views $V_{novel}$, we aim to build a final sample sequence $Sp = [s_1, s_2, ..., s_N]$ where samples from different view sets are evenly interleaved and satisfy:
\begin{equation}
\frac{|S_{ref}|}{|S_{novel}| + |S_{ref}|} = \alpha.    
\end{equation}

\begin{figure}[!t]\centering
  \includegraphics[width=0.9\linewidth]{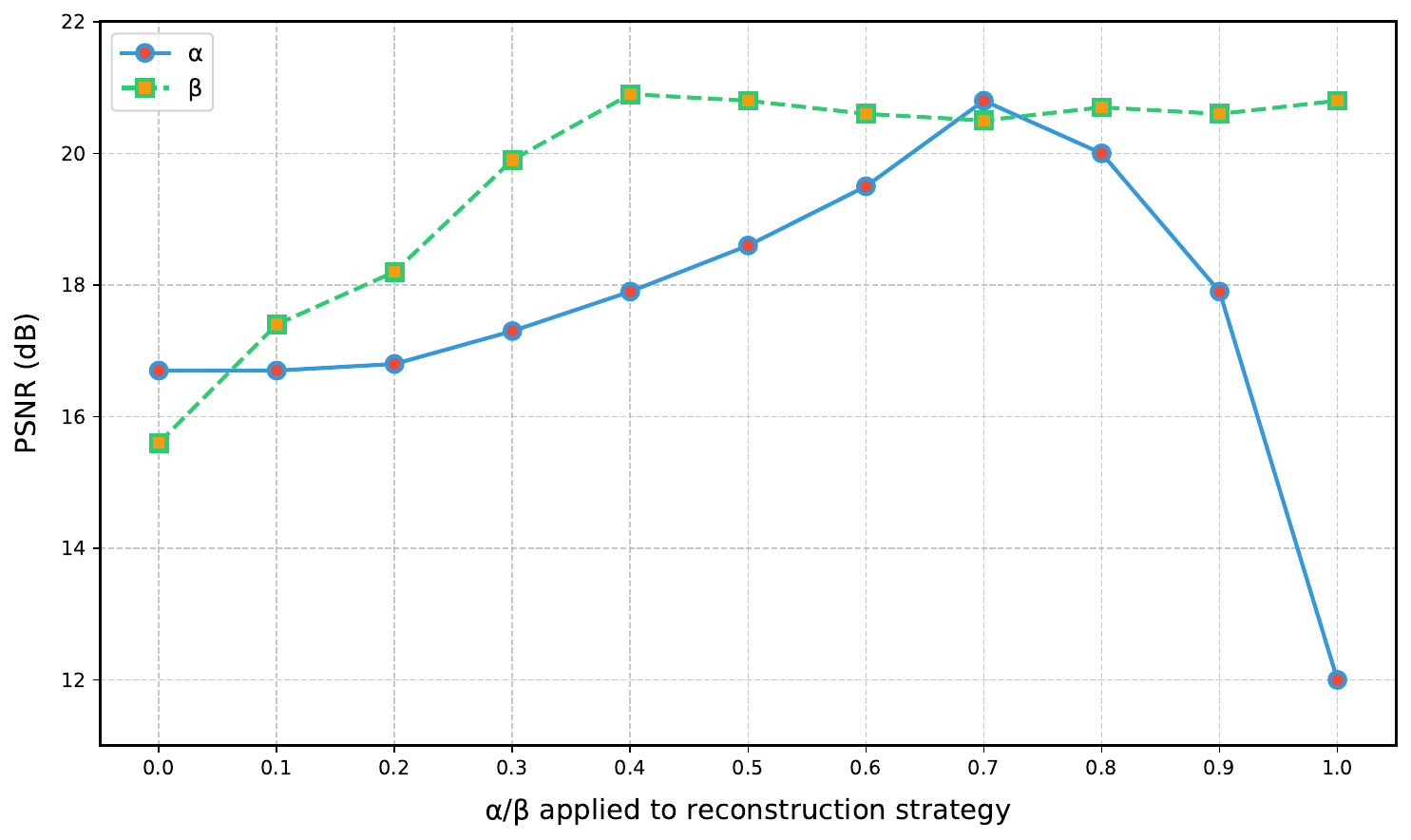}
  \caption{Evaluation on parameter $\alpha$ and $\beta$.}
  \label{fig:alpha}
\end{figure}

Here $S_{ref}$ and $S_{novel}$ refer to samples from $V_{ref}$ and $V_{novel}$, and $\alpha$ is the weight to drop novel samples.
We test the difference of taking different $\alpha$ values, and the results from a set of DL3DV scenes is shown in Figure \ref{fig:alpha}, where $\alpha = 0$ means complete drop on input views and $\alpha = 1$ means complete drop on novel views.
Therefore, in practice, we set $\alpha$ at 0.7.

During training, every drawn sample can be discarded with a dropped probability defined as follows: 
\begin{equation}
P^{drop} = \begin{cases}
    1 - min(1, \frac{\alpha}{r_i}),& if \ s_i \in V_{ref} \\
    1 - min(1, \frac{1-\alpha}{1-r_i}),& if \ s_i \in V_{novel}
\end{cases},
\end{equation}
\begin{equation}
r_i = \frac{|\{s_j|j<i, s_j \in S_{ref}\}|}{|\{s_j|j<i\}|}.
\end{equation}

The sampling strategy generates an interleaved sample sequence and stably controls the proportion of two types of guidance in the training set, providing more stable supervision signals during training.

\begin{algorithm}
\caption{Reconstruction process with random sample drop}\label{alg:rsd}
\SetKwInOut{KwInit}{Initialize}
\SetKwInOut{KwParam}{Parameters}
\KwParam{Iterative view sampling stack $S$, input reference views $V_{ref}$, novel views $V_{novel}$, sample iteration $it_s$, final iteration $it_e$, 3DGS representation $R$, weight $\alpha$ for random sample drop}
\KwOut{well-optimized 3DGS representation $R$}
\KwInit{
$i \gets 0$. \tcp{Iteration}

$r \gets 1e-6$. \tcp{For $P^{drop}$ update.}
}
\For{$i = 1 \ to \ it_s$}{
    Optimize $R$ with reference views.
}
\For{$i = it_s \ to\ it_e$}{
    \If{$S$ is empty}{
        $S$ = shuffle($S_{ref}$ + $S_{novel}$).
    }
    
    Get next sample $s$ from $S$.
    
    Determine whether to drop $s$ according to $P^{drop}(s, \alpha, r)$.
    
    Update $r$.
    
    \If{$not\ to\ drop$}{
        Optimize $R$ with $s$.
    }
    
    Update $i$.
}
\Return{$R$}
\end{algorithm}

\begin{figure*}[!t]\centering
  \includegraphics[width=1\linewidth]{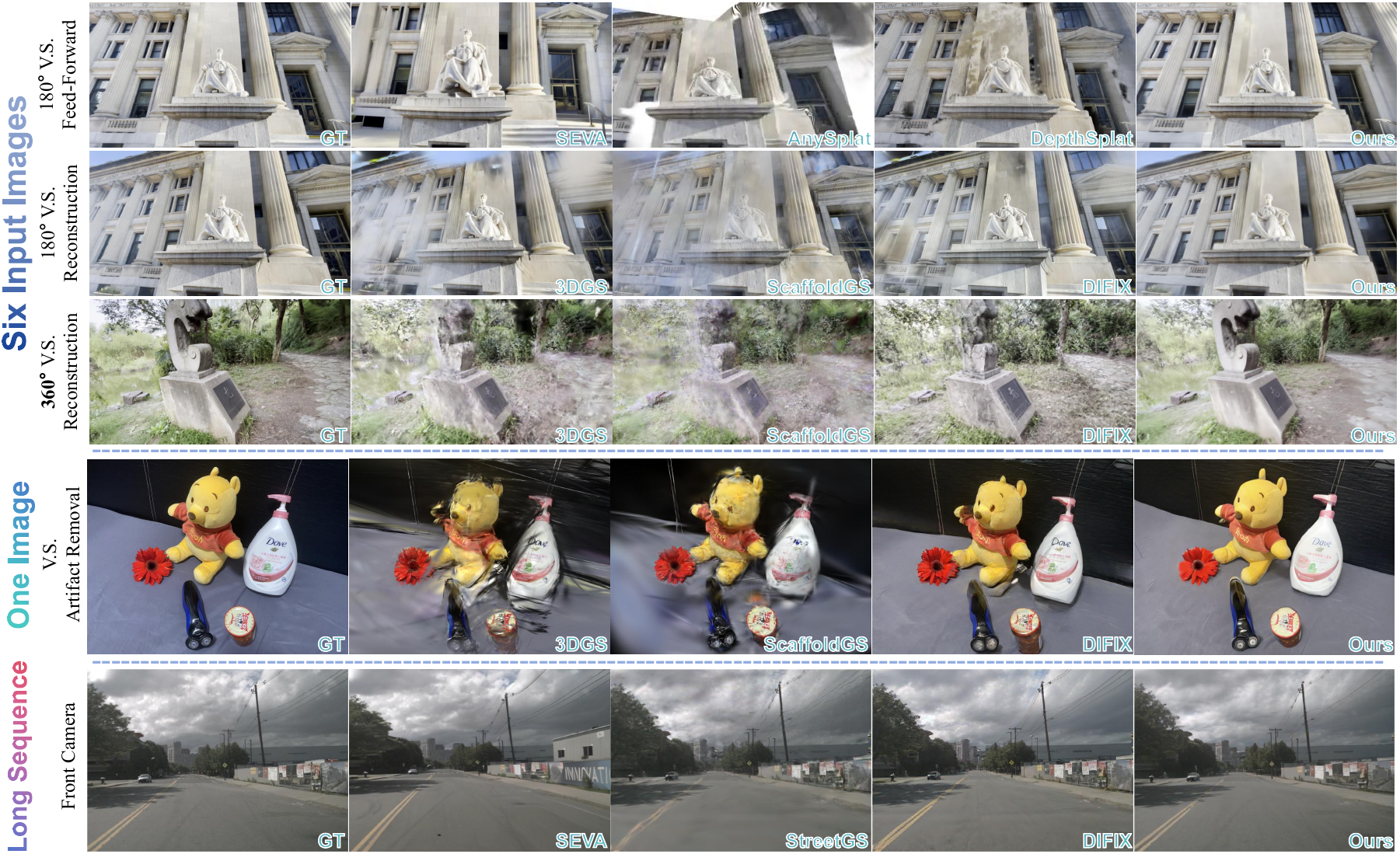}
  \caption{Qualitative results in different situations.}
  \label{fig:qualitative}
\end{figure*}

\textbf{Weighted gradient.}
In terms of large artifacts that severely corrupt novel views, certain bias between ground truth and novel guidance is inevitable.
Previous methods ignore this problem \cite{zhao2025drivedreamer4d, wu2025difix3d+} or adopt a general weight \cite{ni2025recondreamer, zhao2025recondreamer++, yan2025streetcrafter} on novel views, overfitting to incorrect details.
Instead, we design a pixel-level weight $\mathbf{W} \in [0, 1]^{H \times W}$ for each view based on a confidence mask $\mathbf{M}^{conf}$ to reduce the adverse impact of possible inconsistently fixed regions on Gaussian model training:
\begin{equation}
\mathbf{W}_i(x, y) = \begin{cases}
        \beta + (1-\beta)\mathbf{M}^{conf}_{i}(x, y),& s_i \in V_{novel}\\
        1,& s_i\in V_{ref},
\end{cases}
\end{equation}
where weight parameter $\beta$ is also tested as shown in Figure \ref{fig:alpha}.
While $\beta=1$ means no weighted gradient, we set $\beta$ to 0.4 for better artifact suppression and overall quality.
The confidence mask $\mathbf{M}^{conf}$ is calculated by point cloud rendering of a training view $v$:
\begin{equation}
\mathbf{M}^{conf}_{v}(x,y) = \mathbb{I}((x, y) \in \mathcal{P}),
\end{equation}
where $\mathcal{P}$ refers to valid pixels projected by the point cloud, $\mathbb{I}$ refers to indicator function. For a splatted gaussian $g_i$, the gradient $\nabla_{i}L$ based on $\mathbf{W}_i$ is calculated as follows:
$$
\nabla_{i}L = \frac{1}{|\mathcal{P}_i|}\sum_{x,y\in\mathcal{P}_i}\mathbf{W}_i(x,y)\frac{\partial L}{\partial{G_{i}(x,y)}},
$$
where $\mathcal{P}_i$ denotes pixels set $g_i$ splats on, $G_i(x, y)$ denotes the gaussian function value on $(x, y)$. Due to the constraint of $\mathbf{W}_i$ on pixels, regions containing potential artifacts will contribute less in backpropagation process.

While the artifact fixing is reliable with strong guidance from point cloud, regions without any point cloud but with large artifacts can update the corresponding Gaussian parameters more conservatively with the above strategy.
This avoids oscillation or failure of gaussian model by dominating the optimization process with input views and point cloud guidance, providing better 3D consistency.

Ablations of the method can be found in Sec \ref{sec:ablation}.

%% file: sec/4_experiment.tex
\section{Experiments}
We conduct extensive experiments including indoor scenes, outdoor scenes and driving scenes for evaluation, providing both quantitative and qualitative results. 
\subsection{Implementation Details}
Our artifact fixer is trained using the train split of DL3DV-960 \cite{ling2024dl3dv, fcgs2025}, including around 2.1M image pairs.
The processing for paired data curation (detailed in Sec \ref{sec:method_fixer}) took 850 H200 GPU hours,
and the fixing model is trained on 8 H200 GPU for 60 hours with a per-GPU batch size of 4.
Evaluation is conducted on a single RTX 4090 GPU for a more general comparison of user side performance.
Sample iteration $it_s$ is set to $3,000$ for static scene reconstruction, and $7,000$ for driving scene reconstruction.
Final iteration $it_e$ is fixed to $30,000$ for all scenes.
We use 3DGS \cite{kerbl20233d} as reconstruction backbone for S2D and DIFIX, other methods all follow their original settings.
For pure generative methods, structural metrics are not applied because they do not support precise camera control on long sequences with all test cameras included in a single run.
More implementation details can be found in Appendix.

\begin{table*}
\centering
\resizebox{\textwidth}{!}{
\begin{tabular}{lcccccccccccccccc}
\toprule
\multirow{2}{*}{Datasets} & \multicolumn{4}{c}{3DOVS @ 1 Image} & \multicolumn{4}{c}{RE10K @ 2 Images} & \multicolumn{4}{c}{MIP360 @ 6 Images} & \multicolumn{4}{c}{DL3DV @ 6 Images} \\ 
\cmidrule(lr){2-5} \cmidrule(lr){6-9} \cmidrule(lr){10-13} \cmidrule(lr){14-17}
 & PSNR$\uparrow$ & SSIM$\uparrow$ & LPIPS$\downarrow$ & FID$\downarrow$ & PSNR$\uparrow$ & SSIM$\uparrow$ & LPIPS$\downarrow$ & FID$\downarrow$ & PSNR$\uparrow$ & SSIM$\uparrow$ & LPIPS$\downarrow$ & FID$\downarrow$ & PSNR$\uparrow$ & SSIM$\uparrow$ & LPIPS$\downarrow$ & FID$\downarrow$ \\
 \midrule
3DGS\cite{kerbl20233d} & 10.12 & 0.30 & 0.69 & 278.9 & 21.14 & 0.61 & 0.36 & 64.3 & 17.14 & 0.48 & 0.41 & 71.9 & 15.26 & 0.51 & 0.46 & 97.7 \\
Mip-Splatting\cite{Yu2024GOF} & 11.31 & 0.29 & 0.61 & 249.6 & 21.35 & 0.65 & 0.31 & 63.2 & 16.33 & 0.44 & 0.48 & 91.3 & 18.50 & 0.59 & 0.41 & 86.3 \\
ScaffoldGS\cite{scaffoldgs} & 10.50 & 0.23 & 0.62 & 230.4 & 20.96 & 0.60 & 0.32 & 67.6 & 16.02 & 0.35 & 0.46 & 162.2 & 17.41 & 0.60 & 0.39 & 89.7 \\
\midrule
NoPoSplat\cite{ye2024noposplat} & - & - & - & - & 24.30 & 0.70 & 0.36 & 41.0 & 15.48 & 0.37 & 0.46 & 91.4 & 18.11 & 0.50 & 0.41 & 86.4 \\
AnySplat\cite{jiang2025anysplat} & 17.91 & 0.48 & 0.42 & 67.0 & 24.32 & 0.69 & 0.35 & 40.6 & 18.12 & 0.60 & 0.44 & 96.4 & 17.39 & 0.59 & 0.43 & 85.3 \\
DepthSplat\cite{xu2025depthsplat} & - & - & - & - & 25.79 & 0.76 & 0.33 & 38.9 & 19.24 & 0.61 & 0.40 & 84.2 & 19.62 & 0.68 & 0.37 & 81.4 \\
\midrule
ViewCrafter\cite{yu2024viewcrafter} & - & - & 0.36 & 71.2 & - & - & 0.30 & 39.2 & - & - & - & - & - & - & - & - \\
SEVA\cite{zhou2025stable} & - & - & 0.33 & 67.3 & - & - & 0.29 & 37.4 & - & - & 0.36 & 60.2 & - & - & 0.30 & 52.8 \\
\midrule
DIFIX\cite{wu2025difix3d+} & 14.10 & 0.47 & 0.56 & 98.2 & 26.11 & 0.80 & 0.28 & 40.2 & 19.43 & 0.60 & 0.38 & 70.3 & 20.4 & 0.72 & 0.32 & 55.3 \\
S2D (Ours) & \textbf{21.41} & \textbf{0.77} & \textbf{0.27} & \textbf{54.0} & \textbf{27.62} & \textbf{0.86} & \textbf{0.24} & \textbf{34.7} & \textbf{20.97} & \textbf{0.70} & \textbf{0.35} & \textbf{58.1} & \textbf{23.2} & \textbf{0.81} & \textbf{0.26} & \textbf{41.2} \\
\bottomrule
\end{tabular}
}
\caption{Quantitative results in in-the-wild scenes.}
\label{tab:all}
\end{table*}

\subsection{Datasets and Metrics}
For indoor and outdoor scene reconstruction, we use 3DOVS \cite{liu2023weakly}, MIP360 \cite{barron2022mipnerf360} and the test split of DL3DV-960 \cite{ling2024dl3dv, fcgs2025} and RE10K \cite{zhou2018stereomagnificationlearningview}.
For driving scene artifact removal and reconstruction, we use Waymo Open Dataset \cite{sun2020scalability}.

To evaluate the structural quality of scene reconstruction, we use the peak signal-to-noise ratio (PSNR) and similarity index measure (SSIM).  
We also report perceptual distance (LPIPS) \cite{zhang2018unreasonable} and fréchet inception distance (FID) \cite{heusel2017gans} for perceptual quality, especially in driving scenes with no ground truth provided in extrapolated trajectories.

\subsection{Comparison on in-the-wild scenes}
We compare the reconstruction quality of our method with baseline 3DGS, feed-forward methods and also generative methods.
For front-facing scenes in 3DOVS, we use only one image as input.
For other datasets with larger camera movements, we use six images as input.

\textbf{Quantitative results.}
The quantitative results are reported in Table \ref{tab:all}.
Under such extreme inputs, both traditional methods and state-of-the-art feed-forward methods show significant degradation.
Generative methods shows good ability in perceptual quality, but cannot consistently support long sequence with precise camera control, which is not suitable for explicit 3D reconstruction of large scenes.
S2D clearly outperforms the above methods, and also shows much improvement comparing with sparse-view reconstruction solution DIFIX.

\textbf{Qualitative results.}
The qualitative results are shown in Figure \ref{fig:qualitative} for more visual comparison.
For $180^\circ$ view range with six input images, generative method SEVA \cite{zhou2025stable} provides inconsistent details (\eg the statue and extra pole to the left), and feed-forward methods \cite{jiang2025anysplat, xu2025depthsplat} either suffer from limited view range or flying artifacts.
While traditional methods \cite{kerbl20233d, scaffoldgs} inevitably face large artifacts, the scene reconstructed using DIFIX \cite{wu2025difix3d+} enhancement still can't avoid them, which degrades more in $360^\circ$ view range.
Only our method provides the most stable and clean reconstruction under inputs with different sparsity.

Meanwhile, we also showcase the artifact removal ability under another demanding input settings.
With only one input, the 3DGS result not only shows severe artifacts, but also suffers from position jitter.
Operating on the same 3DGS artifact image, the result of DIFIX is quite blurry and the position can not be fixed, while our result is correct and neat.
All the above results show that our pipeline is the strongest in terms of sparse inputs.

\subsection{Comparison on driving scenes}
In driving scenes, although the input sequence is long, both the large capture distance and the nature of single trajectory could cause significant artifacts in reconstructed scenes, where a solution is also being expected.
In this case, we also evaluate our method's ability when applied to driving scene reconstruction.

\textbf{Quantitative results.}
For view interpolation on input trajectory, we compare with base driving scene reconstruction methods StreetGaussians \cite{yan2024street} and EmerNeRF \cite{yang2023emernerf}.
For view extrapolation with lane shifts, we mainly compare with StreetCrafter \cite{yan2025streetcrafter} and DIFIX \cite{wu2025difix3d+}.
The shift setting is similar to StreetCrafter with side shift (left and right) and up shift.

The results are provided in Table \ref{tab:waymo}, where base reconstruction methods shows extremely high FID in shifted views due to large artifacts.
While DIFIX can fix small artifacts to improve quality on interpolated views, its lane shift FID is higher than StreetCrafter which uses a video generation model trained on driving scenes for novel trajectory generation.
In contrast, our method shows good capability in driving scene reconstruction and outperforms previous methods on both view interpolation and extrapolation.

\textbf{Qualitative results.}
We also provide qualitative results in Figure \ref{fig:qualitative} Bottom and Appendix.
In Figure \ref{fig:qualitative} is comparison of interpolated front camera views from NuScenes \cite{caesar2020nuscenes} dataset.
The direct video generation from SEVA \cite{zhou2025stable} creates inconsistent background objects.
Operating on the artifact rendering from StreetGS, the image fixed by DIFIX has clear deformation on lane lines, and the road is also less smooth comparing with our method.
More comparison can be found in Appendix.

\begin{table}
\centering
\resizebox{\columnwidth}{!}{
\begin{tabular}{lccccc}
\toprule
\multirow{2}{*}{Methods} & \multicolumn{2}{c}{Interpolation} & \multicolumn{3}{c}{Lane Shift @ FID$\downarrow$} \\ 
\cmidrule(lr){2-3} \cmidrule(lr){4-6} 
 & PSNR$\uparrow$ & LPIPS$\downarrow$ & side 2m & side 3m & up 1.5m \\
 \midrule
EmerNeRF\cite{yang2023emernerf} & 25.32 & 0.20 & 96.4 & 101.3 & 90.2 \\
StreetGaussians\cite{yan2024street} & 30.14 & 0.14 & 75.0 & 96.7 & 70.3 \\
StreetCrafter\cite{yan2025streetcrafter} & 29.31 & 0.10 & 57.4 & 66.4 & 59.0 \\ 
DIFIX\cite{wu2025difix3d+} & 30.26 & 0.11 & 60.2 & 71.6 & 60.3 \\
S2D (Ours) & \textbf{31.44} & \textbf{0.07} & \textbf{46.1} & \textbf{53.9} & \textbf{41.3} \\
\bottomrule
\end{tabular}
}
\caption{Quantitative results on Waymo Open Dataset.}
\label{tab:waymo}
\end{table}

%% file: sec/5_ablation.tex
\section{Ablation Studies}
We demonstrate the rationality of our design by conduct ablation studies.
\label{sec:ablation}
\begin{figure*}[t]\centering
  \includegraphics[width=1\textwidth]{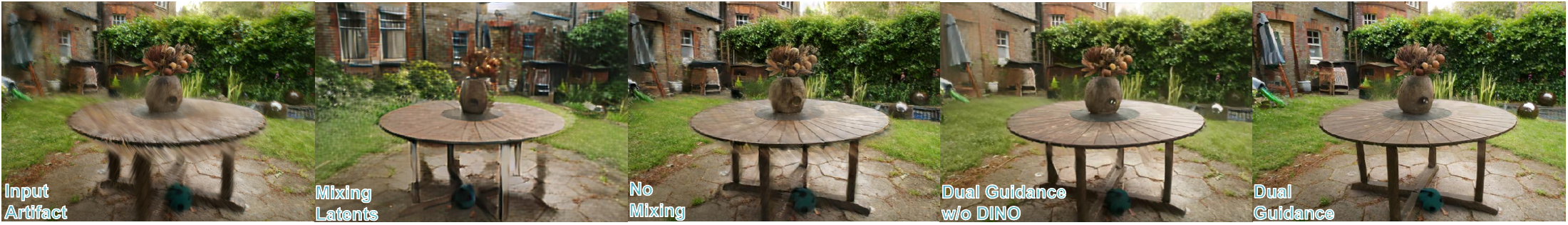}
  \caption{Comparison of artifact removal with ablations on guidance mixing.}
  \label{fig:ablation_mixing}
\end{figure*}
\subsection{Guidance Mixing}
\label{sec:guidancemixing}
\begin{table}[t]
\centering
\caption{Ablation on dual guidance in artifact fixer.}
\label{tab:ablation_dualguidance}
\resizebox{\linewidth}{!}{
\begin{tabular}{c|lcc}
  \toprule
  \# & Ablations & PSNR $\uparrow$ & LPIPS $\downarrow$\\
  \midrule
  \ding{172} & Generation from point cloud & 12.7 & 0.71\\
  \ding{173} & Mixing latents & 15.6 & 0.49\\
  \ding{174} & No mixing & 19.0 & 0.38\\
  \ding{175} & Dual guidance w/o DINO & 22.1 & 0.27\\
  \ding{176} & Dual guidance w DINO & 23.0 & 0.26\\
  \bottomrule
\end{tabular}
}
\end{table}
%
We first conduct ablations on the artifact fixer by analyzing current dual guidance design.
The first comparison is to use only point cloud rendering and reference view for fixed view generation.
The idea is to utilize both structure information from point cloud and background information from reference view, but the model fails to recover accurate background in a single denoising step (\ding{172}).

When removing the mixing module for the input view, we directly batch point cloud rendering and reference view together with artifact image through multi-view UNet as in DIFIX \cite{wu2025difix3d+}, where we found the model tends to overlook the point cloud input and uses only other images (\ding{174}),
while using DINO guidance will help with balanced attention during the mixing, which is useful to some extent but not significant (\ding{175} \& \ding{176}).
We believe this is because the early loss is mainly dominated by general image quality, thus the detailed structural guidance of point cloud is covered up.

We also test putting mixing module after VAE block which directly operates upon encoded latents, 
and we found the results to be overly smoothed (\ding{173}).
Such operation on latents may have caused the targets to drift away from the original space determined by denoising backbone.

The statistical and qualitative ablation results are shown in Table \ref{tab:ablation_dualguidance} and Figure \ref{fig:ablation_mixing}, which prove the rationality of current artifact fixer design.

\subsection{Reconstruction Strategy}
\label{sec:reconstrategy}
\begin{figure}[!t]\centering
  \includegraphics[width=1\linewidth]{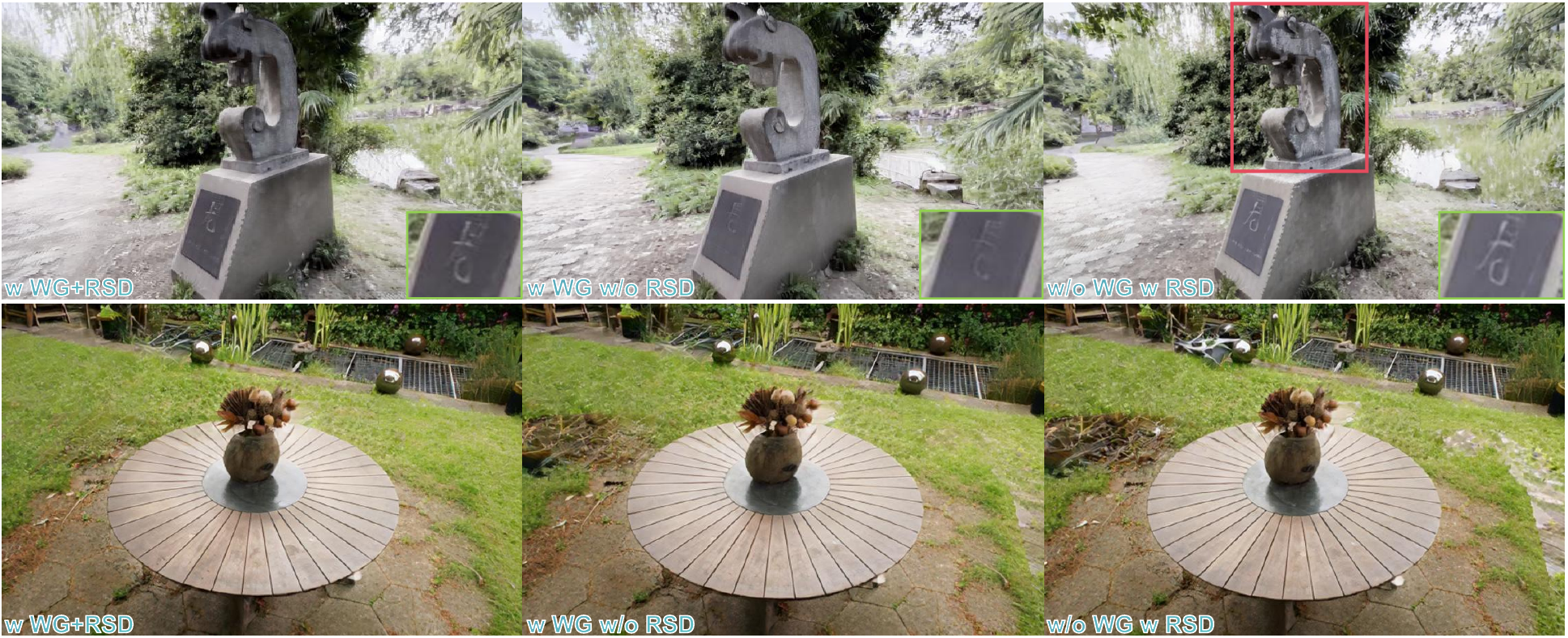}
  \caption{Ablations on reconstruction strategy. WG indicates weighted gradient, and RSD for random sample drop.}
  \label{fig:ablation}
\end{figure}
We also conduct ablations on our reconstruction strategy, which is important in consistent 3D lifting.

Small inconsistency is inevitable with large view range, especially in areas unseen from limited input views.
In this case, random sample drop first ensure stable presence of input views, and weighted gradient further constrain the update of wrong areas.

The iterative training stack maintained by random sample drop can provide sufficient 3D correct guidance from input views.
Without this behavior, structural details such as textures or small texts (which is a common difficulty for generative methods) will be averaged by novel guidance, as shown in Figure \ref{fig:ablation} (line 1 middle).
 
For larger inconsistency occurred in severe artifacts, weighted gradient plays a bigger role.
As shown in Figure \ref{fig:ablation}, the large artifacts can not be simply resolved by more samples from input views, and weighted gradient is important in stopping the permanent update of corrupted areas that are also missing in point cloud guidance.

In general, the above ablations prove the necessity of our design for artifact fixer and reconstruction strategy.

For more discussion and demonstration, please refer to Supplementary Material.

%% file: sec/6_conclusion.tex
\section{Conclusion}
In this paper, we present Sparse to Dense lifting (S2D), a flexible novel pipeline that enables 3DGS reconstruction with minimal inputs.
We present a strong and efficient diffusion fixer for high fidelity artifact fixing along with corresponding reconstruction strategy for robust model fitting.

Experiments shows that our method provides the first-tier results, while the rationality of core components are proven with comprehensive ablation studies.
S2D can be easily applied to most 3DGS methods to largely reduce their input requirements, encouraging more general 3DGS applications in real-world situations.

%% file: sec/7_ack.tex
\section*{Acknowledgements}
This work was supported by the National Natural Science Foundation of China (Grant
Nos. 72192821, 62302297, 62302167, 62472282, 62222602, 62502159, U23A20343, W2521174), the Shanghai Committee of Science and Technology (Grant Nos. 25511103300, 25511104302, 25511102700), the Fundamental Research Funds for the Central Universities (project number: YG2023QNA35), YuCaiKe [2023] Project Number: 231111310300, and the Young Elite Scientists Sponsorship Program by CAST YESS20240780.

%% file: sec/X_suppl.tex
\clearpage
\setcounter{page}{1}
\maketitlesupplementary

\section{Ablation on loss terms}
\begin{table}
\centering
\resizebox{\linewidth}{!}{
\begin{tabular}{lcc}
  \toprule
  Ablations & PSNR $\uparrow$ & LPIPS $\downarrow$\\
  \midrule
$\mathcal{L}_{LPIPS}+\mathcal{L}_{2}+\mathcal{L}_{GAN}+\mathcal{L}_{CLIP}$  & 20.1 & 0.35\\
$\mathcal{L}_{LPIPS}+\mathcal{L}_{2}+\mathcal{L}_{GAN}$  &  20.1 & 0.35\\
$\mathcal{L}_{LPIPS}+\mathcal{L}_{2}+\mathcal{L}_{GAN}+\mathcal{L}_{DINO}$  &  20.5 & 0.33\\
$\mathcal{L}_{LPIPS}+\mathcal{L}_{2}+\mathcal{L}_{GAN}+\mathcal{L}_{SSIM}$ & 21.8 & 0.28\\
$\mathcal{L}_{LPIPS}+\mathcal{L}_{2}+\mathcal{L}_{GAN}+\mathcal{L}_{SSIM}+\mathcal{L}_{DINO}$ & 22.0 & 0.27\\
  \bottomrule
\end{tabular}}
\caption{Ablation on loss terms.}
\label{tab:ablation_loss}
\end{table}

We further conduct ablations on loss terms applied on S2D artifact fixer by evaluating the image quality conditioned on different loss choices.
Apart from base losses that are already verified \cite{parmar2024one}, we compare the difference of CLIP, DINO and SSIM losses.
The results are reported in Table \ref{tab:ablation_loss}.

While this is not new image generation task, the original CLIP loss actually contributes nothing in training, thus removing CLIP loss makes completely no difference.
Introducing SSIM loss provides more improvements with higher supervision on structural details.
DINO loss also helps with small enhancement by aligning semantic features.

We also test applying different weights to each loss term, and the results remain unchanged across reasonable variation within the range of 0.4.

\section{Training data}
\label{sec:appendix_data}
\begin{figure}[!t]\centering
  \includegraphics[width=1\linewidth]{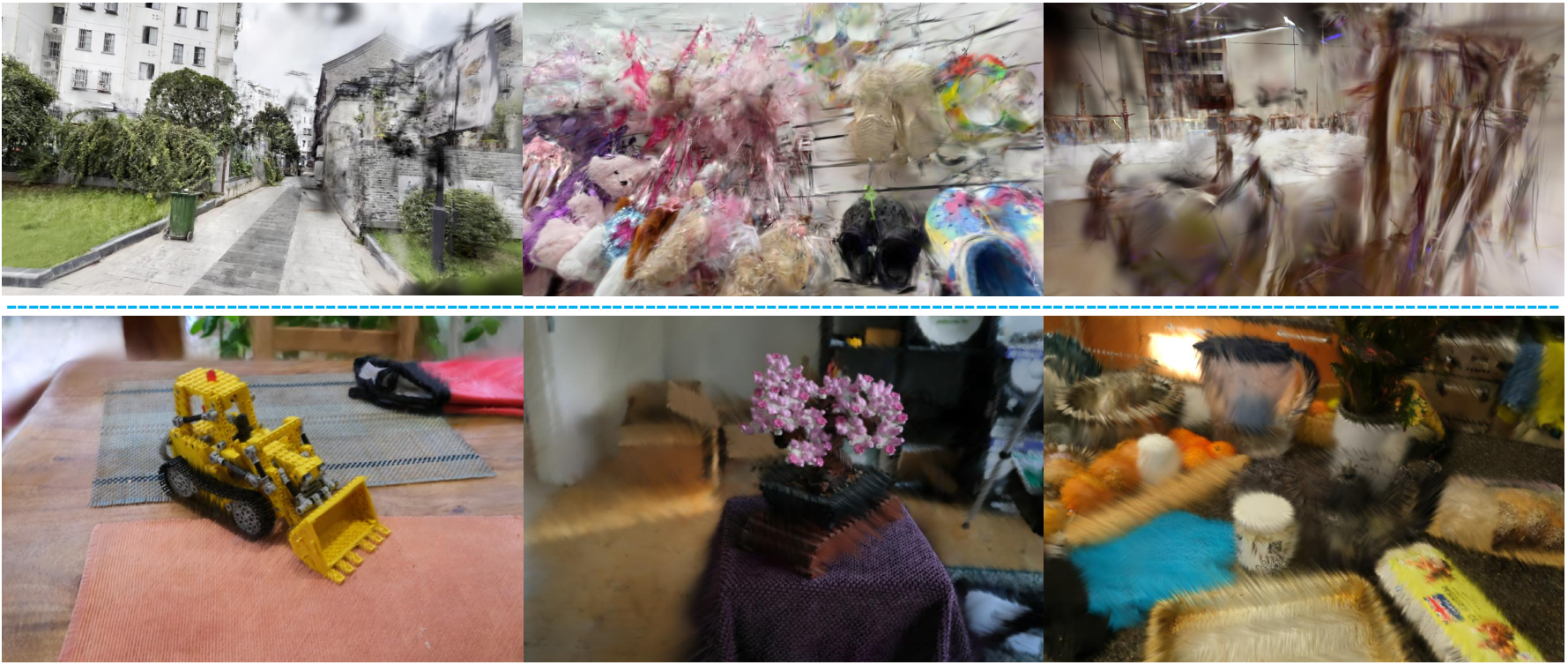}
  \caption{Training data with different levels of artifacts (line 1) and example of random perturbations (line 2).}
  \label{fig:disturb}
\end{figure}

To enhance the robustness of our model, we generate an extensive dataset that covers the broadest spectrum of corruption boundaries.

The processed training data with different artifact levels is shown in Figure \ref{fig:disturb} Line 1.
To demonstrate how random perturbations introduced in Sec \ref{sec:method_fixer} works, we also provide an example of render-time perturbation on correctly reconstructed 3DGS scenes in Figure \ref{fig:disturb} Line 2, where the perturbation intensity increases from left to right.

The large artifacts guide the model to recover basic object structure, while the perturbation applied on mild corruptions can further encourage detail enhancements, especially textures and edges.
Therefore, our model is able to handle a large range of input intensity.

\section{Evaluation on input density}
\begin{figure}[!t]\centering
  \includegraphics[width=1\linewidth]{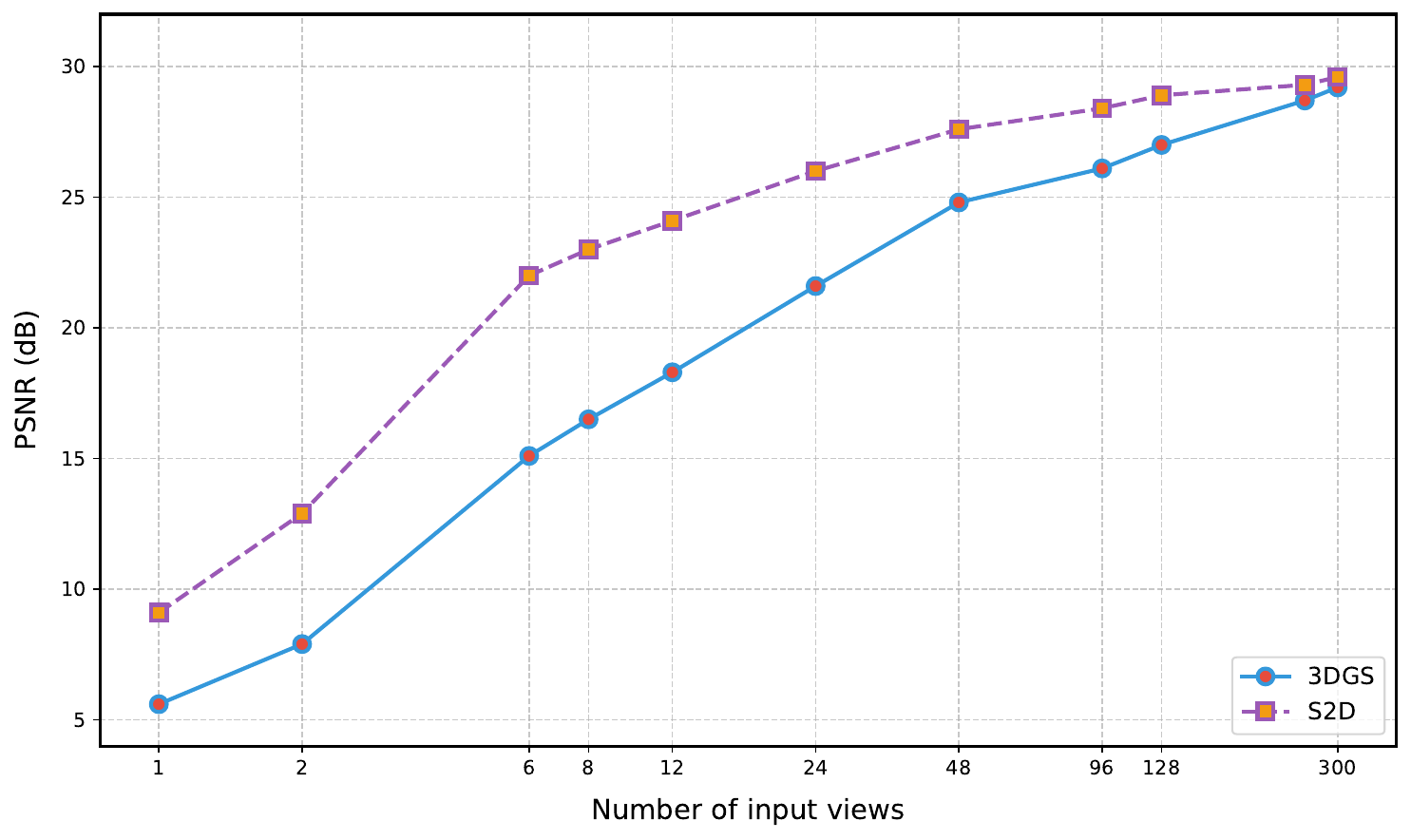}
  \caption{Reconstruction quality versus input density.}
  \label{fig:psnr_vs_density}
\end{figure}

We evaluate S2D's ability when faced with different amount of input images, the results on DL3DV scenes are provided in Figure \ref{fig:psnr_vs_density}.

As shown, our reconstruction quality improves steadily when increasing the number of input views.
S2D consistently outperforms 3DGS with large improvements in sparse input and extra enhancement in dense input settings.
This shows that S2D can be applied in general situations instead of fixed input views.

\section{Performance report}
The computational cost of S2D fixer is low, including 11.1 GB GPU usage and 1 FPS processing speed upon image resolution $1024\times576$ on a single RTX 4090.
Under the same setting, Stable Virtual Camera (SEVA) results in 16.4 GB GPU usage and 0.08 FPS generation speed, while DIFIX is on par with S2D with 10.9 GPU usage and 1 FPS processing speed.

The reported processing speed means that the generation of novel guidance with S2D creates very small overhead comparing with original reconstruction (\eg on 3DOVS dataset, with 30 seconds fixing is only $1/30$ of the total 15 minutes reconstruction), empowering the application of S2D in diverse scheme combinations while eliminating concerns about noticeable efficiency degradation.
\begin{figure*}[!t]\centering
  \includegraphics[width=0.9\linewidth]{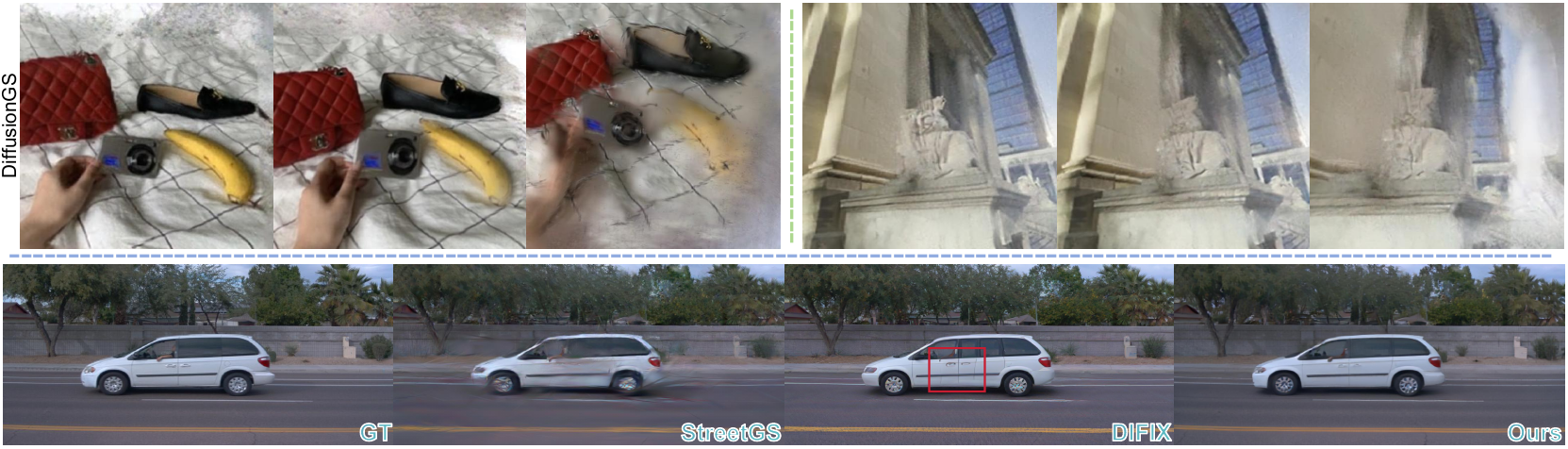}
  \caption{Extra results of DiffusionGS and driving scene comparison.}
  \label{fig:appendix_diffusiongs}
\end{figure*}
\section{More Comparison}

We provide more qualitative comparison in Figure \ref{fig:appendix_diffusiongs}.
For the same scenes as in Figure \ref{fig:qualitative}, the latest sparse-view reconstruction method DiffusionGS \cite{diffusiongs} (mainly object-centric) largely degrades with respect to in-the-wild scenes.

In line 2 we compare the side camera results on Waymo Open Dataset \cite{sun2020scalability}.
Operating on the same artifact image, DIFIX generates wrong car details and fuzzy road with supersaturated lane lines, while ours is more consistent.

\section{More Implementation Details}
Due to the page limit of the main paper, we provide more implementation details and discussion here.

\textbf{Evaluation standard.}
While many feed-forward reconstruction researches now tend to conduct evaluation only on a few images from the original scene capture (especially sparse-view reconstruction such as 2-view reconstruction), we believe that this is insufficient to represent the actual and overall reconstruction quality of the complete scene.

Therefore, evaluation results reported in this paper are tested on \textbf{all other frames} of a scene that are not used for training or reference, identical to the evaluation standard of traditional reconstruction methods.

Evaluation on static scenes is conducted with the max pixel size of $255000$ (the original setting for $\pi^3$ \cite{wang2025pi3}, \eg $672\times378$) per image while remaining the original aspect ratio.
This is to align with the default rendered image of point cloud for detail consistency, which is not compulsory.
While many feed-forward methods only support fixed resolution or aspect ratio such as $256\times256$, S2D can work on any resolution and aspect ratio.
In driving scenes, we evaluate the results on resolution $1024\times576$ as the setting in StreetCrafter \cite{yan2025streetcrafter}.

\textbf{Rendering for point cloud.}
We render point cloud images generated by VFMs through \textit{pyrender}.

Specifically, we initialize an empty scene with black background, and generate mesh from reconstructed point cloud.

We applied \textit{MetallicRoughnessMaterial} with \textit{metallicFactor} set at 1.0 and \textit{baseColorFactor} set as [1.0, 1.0, 1.0, 1.0] to all meshes to maintain the original point color.

Given cameara intrinsics and poses, the corresponding point cloud images are then rendered through \textit{PerspectiveCamera}.

\textbf{Data and training.}
The S2D artifact fixer is trained on our processed DL3DV-960\cite{ling2024dl3dv, fcgs2025} dataset (train split) on 1-7K scenes for 75000 steps.
The training images are augmented with random horizontal flip.
All images are resized to $512\times512$ during training.
The model is implemented in PyTorch and optimized with the AdamW optimizer, weight dtype is float32.

\section{Limitation and future work}
\begin{figure}[!t]\centering
  \includegraphics[width=1\linewidth]{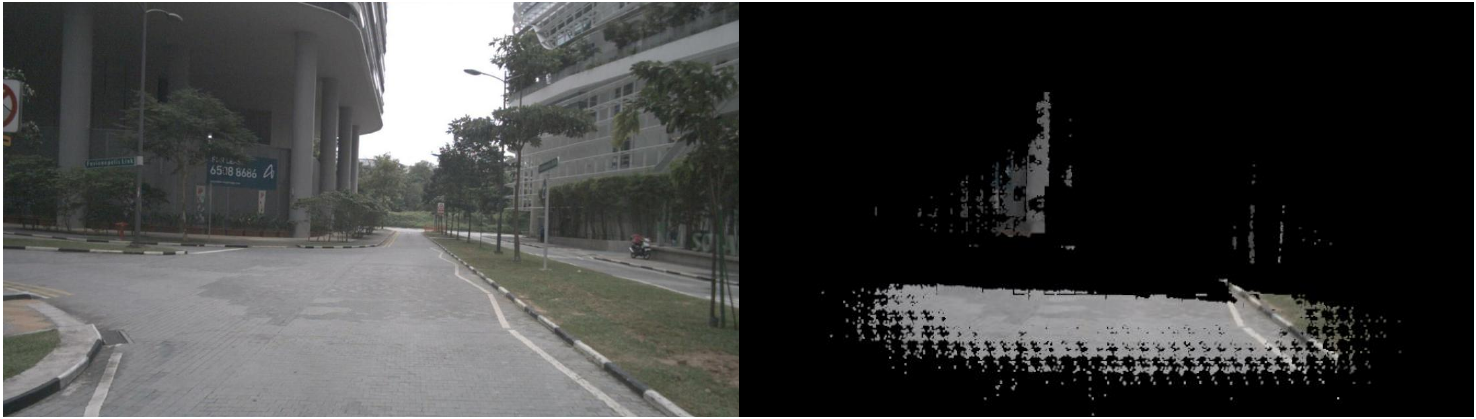}
  \caption{Point cloud rendering with single image input.}
  \label{fig:appendix_failure_case}
\end{figure}

While our methods utilize VFM priors, we share the same limitation as VFMs.
If the input images are both extremely sparse and low in texture, the resulting point cloud can be quite fragmentary.
We show such a failure case in Figure \ref{fig:appendix_failure_case} from the NuScenes dataset.
This means the point cloud can hardly provide sufficient guidance and the fixing can only work on smaller artifacts.
Fortunately, the VFM can be easily replaced to avoid similar situation.

The current fixing model uses point cloud for structural guidance and reference view for textural guidance, but we think the ultimate goal is to directly utilize both structural guidance and textural guidance from reference views while still work as an efficient image-level fixing.
We are now working on further spatial feature extraction and cross-attention to pursue this objective.